\documentclass[acmtog]{acmart}
\acmSubmissionID{}

\usepackage{booktabs} 

\usepackage{amsmath}
\usepackage{microtype}
\usepackage{soul}
\usepackage{xspace}
\usepackage[ruled]{algorithm2e} 

\SetAlFnt{\small}
\SetAlCapFnt{\small}
\SetAlCapNameFnt{\small}
\SetAlCapHSkip{0pt}

\DeclareMathOperator*{\argmin}{arg\,min}

\newcommand{\name}{Video2StyleGAN\xspace}
\newcommand{\gen}{\mathcal{G}}

\acmJournal{TOG}

\renewcommand\footnotetextcopyrightpermission[1]{}
\begin{document}

\title{\name: Disentangling Local and Global Variations in a Video}

\author{Rameen Abdal}
\affiliation{%
\institution{KAUST}
\country{ Saudi Arabia}}
\email{rameen.abdal@kaust.edu.sa}

\author{Peihao Zhu}
\affiliation{%
\institution{KAUST} 
\country{ Saudi Arabia}}
\email{peihao.zhu@kaust.edu.sa}

\author{Niloy J. Mitra}
\affiliation{ 
\institution{UCL and Adobe Research}
\country{UK}}
\email{nimitra@adobe.com}

\author{Peter Wonka}
\affiliation{ 
\institution{KAUST}
\country{ Saudi Arabia}}
\email{pwonka@gmail.com}

\fancyfoot{}

\begin{abstract}

Image editing using a pretrained StyleGAN generator has emerged as a powerful paradigm for facial editing, providing disentangled controls over age, expression, illumination, etc. However, the approach cannot be directly adopted for video manipulations. We hypothesize that the main missing ingredient is the lack of fine-grained and disentangled control over face location, face pose, and local facial expressions. In this work, we demonstrate that such a fine-grained control is indeed achievable using pretrained StyleGAN by working across multiple (latent) spaces (namely, the positional space, the W+ space, and the S space)
and combining the optimization results across the multiple spaces. Building on this enabling component, we introduce \name that takes a target image and driving video(s) to reenact the local and global locations and expressions from the driving video in the identity of the target image. 
We evaluate the effectiveness of our method over multiple challenging scenarios and demonstrate clear improvements over alternative approaches. 

\end{abstract}


%
%


\maketitle

\thispagestyle{empty}

\section{Introduction}

Generative modeling has seen tremendous progress in recent years. Currently, there are multiple powerful frameworks competing in this area, including generative adversarial networks (GANs)~\cite{Karras2020ada,karras2021aliasfree}, variational autoencoders (VAEs)~\cite{VQVAE2}, diffusion network~\cite{DallE2},  and auto-regressive models (ARs)~\cite{taming}.

In this paper, we focus on GANs and in particular the StyleGAN architecture. This architecture has started a wave of research exploring semantic image-editing frameworks~\cite{shen2020interfacegan,patashnik2021styleclip,abdal2020styleflow}. These frameworks first embed a given photograph into the latent space of StyleGAN and then manipulate the image using latent space operations. Example editing operations in the context of human faces are global parameteric image edits to change the pose, age, gender, or lighting, or style transfer operations to convert images to cartoons of a particular style. While these edits are generally successful, it is still an open challenge to obtain fine-grained control over a given face, e.g., face location in the image, head pose, and the facial expression. While such fine-grained control is beneficial but optional for editing single images, they are an essential building block for creating a video from a single image and other video editing applications.

In our work, we set out to address the following research questions: \textit{How can we embed a given video into the StyleGAN latent space to obtain a meaningful and disentangled representation of the video in latent space? How can we create a video from a single image, mainly by transferring pose and expression information from other videos? }


\begin{figure}[t]
    \centering
    \includegraphics[width=\linewidth]{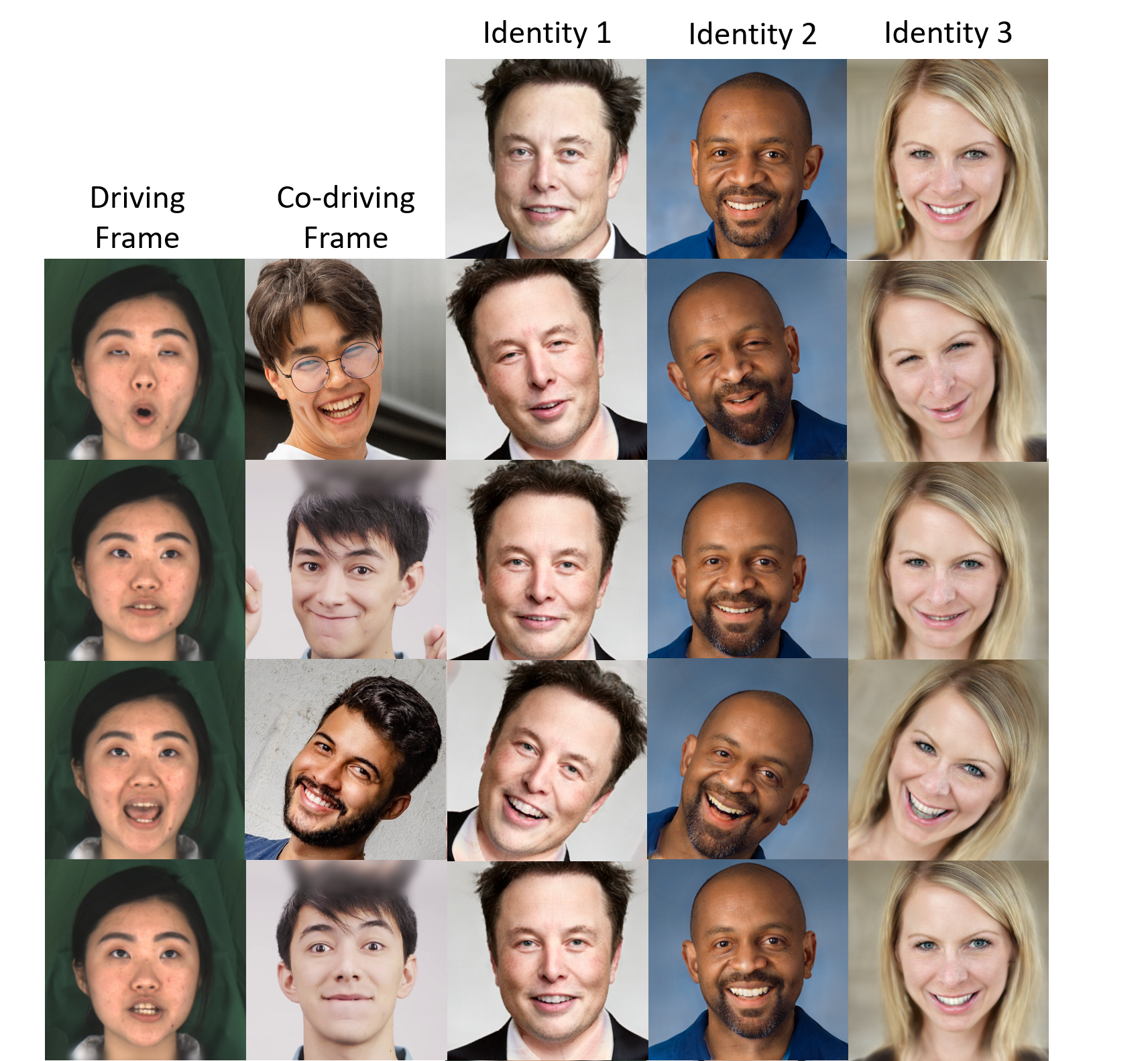}
    \caption{
    \textbf{Fine-grained control.} 
    We present Video2StyleGAN, a video editing framework capable of generating videos from a single image. Our framework can take a driving video and transfer its global and local information to a reference image. We build upon the StyleGAN3 architecture to edit the rotation, translation, pose and local expressions independently so that the information can also be derived from multiple videos.   
    }
    \label{fig:teaser}
\end{figure}

It is actually somewhat surprising how difficult it is to embed fine grained controls into StyleGAN. However, all straightforward solutions are either over-regularized or under-regularized. Over-regularization leads to the controls being ignored so that the given reference image hardly changes. Under-regularization leads to very unnatural face deformations and identity loss.
The main idea of our solution is to make use of different latent spaces to encode different types of information: \textit{positional code} controls the location of the face in the image (i.e. translation and rotation); \textit{W space} controls global edits such as pose and some types of motion; \textit{S space} and generator weights control local and more detailed edits of facial expressions.
This hierarchical (code) structure allows the extraction of semantic information from given driving videos and their transfer to a given photograph. See Figure~\ref{fig:teaser}. 

We compare to a video processing baseline method that does not use multi-resolution latent spaces, as done in recent unpublished arXiv papers~\cite{DBLP:journals/corr/abs-2201-13433,https://doi.org/10.48550/arxiv.2201.08361}. 

The contributions of this work are as follows:
\begin{enumerate}
    \item We propose a facial reenactment system that uses the pretrained StyleGAN3 to transfer the motion and local movements of a talking head that leads to temporally consistent video editing without the requirement of additional training on videos. Our method can transfer these facial movements from a source video onto a single (target) image.   
    \item Our insights into the $W$ and the $S$ space of StyleGAN3 allow us to disentangle both local and global variations in a video. Particularly, we can separately control the eye, nose, and mouth movements. Additionally, we can modify the pose, rotation and translation parameters in a video without compromising the identity of a person.
    \item Finally, in contrast to previous works that take a single video as an input to modify the facial attributes of a given image, we directly extract the local and global variations from multiple videos to reenact a given image. For example, we are able to modify local features like eye, nose, and mouth variations from one video, and other global features like pose, rotation, and translation derived from another video. 
\end{enumerate}

\section{Related Work}

\subsection{State-of-the-art GANs}
Recent improvements to the loss functions, architecture and availability of high quality datasets~\cite{STYLEGAN2018} have improved the generation quality and diversity of Generative adversarial Networks (GANs)~\cite{goodfellow2014generative,radford2015unsupervised}. Owning to these developments, Karras et. al published a sequence of architectures~\cite{karras2017progressive,STYLEGAN2018,Karras2019stylegan2,Karras2020ada,karras2021aliasfree} leading to state-of-the-art results on high quality datasets like  FFHQ~\cite{STYLEGAN2018}, AFHQ~\cite{choi2020starganv2} and LSUN objects~\cite{yu15lsun}. The latent space learned by these GANs have been explored to perform various tasks such as image editing~\cite{shen2020interfacegan,abdal2019image2stylegan,patashnik2021styleclip,abdal2020styleflow} or unsupervised dense correspondence computation~\cite{peebles2021gansupervised}. While recent 3D GANs showed promise in generating high-resolution multi-view-consistent images along with approximate 3D geometry~\cite{chan2021efficient,deng2021gram,orel2021stylesdf}, their quality still lags behind 2D GANs and code for the best methods is not yet available. In this work, we build upon the state-of-the-art generator StyleGAN3~\cite{karras2021aliasfree}. StyelGAN3 exhibits nice properties of translation and rotation in-variance with respect to the generated image. These properties attract research into disentangled global and local video editing.

\subsection{Image Projection and Editing using GANs}

There are two building blocks required for GAN-based image and video editing. First, one needs to project real images into the GAN's latent space. In the StyleGAN domain, Image2StyleGAN~\cite{abdal2019image2stylegan} uses the extended $W+$ latent-space to project a real image into the StyleGAN latent space using optimization. Focusing on improving the reconstruction-editing quality trade-off, other methods including II2S~\cite{zhu2020improved} and PIE~\cite{tewari2020pie} propose additional regularizers to make sure the optimization converges to a high density region in the latent space. While other works like IDInvert~\cite{zhu2020domain}, pSp~\cite{richardson2020encoding}, e4e~\cite{tov2021designing}, and Restyle~\cite{alaluf2021restyle} use encoders and identity preserving loss functions to maintain the semantic meaning of the embedding. Two recent works, PTI~\cite{roich2021pivotal} and HyperStyle~\cite{DBLP:journals/corr/abs-2111-15666} modify the generator weights via an optimization process and Hyper network respectively. Such methods improve the reconstruction quality of the projected images.\par 
Second, latent codes need to be manipulated to achieve a desired edit. For the StyleGAN architecutre, InterFaceGAN~\cite{shen2020interfacegan}, GANSpace~\cite{harkonen2020ganspace}, StyleFlow~\cite{abdal2020styleflow}, and StyleRig~\cite{tewari2020stylerig} propose linear and non-linear edits of the underlying $W$ and $W+$ spaces. StyleSpace~\cite{wu2020stylespace} argues that the $S$ space of StyleGAN leads to better edits. CLIP~\cite{DBLP:journals/corr/abs-2103-00020} based image editing~\cite{patashnik2021styleclip,gal2021stylegan,cl2sg} and domain transfer~\cite{zhu2022mind,DBLP:journals/corr/abs-2112-11641} also study the StyleGAN and CLIP latent spaces to apply StyleGAN based editing on diverse tasks.

\vspace{-0.3cm}
\subsection{GAN based Video Generation and Editing}
GAN based video generation and editing methods~\cite{DBLP:journals/corr/abs-2101-12195,https://doi.org/10.48550/arxiv.2004.01823, Tulyakov:2018:MoCoGAN,https://doi.org/10.48550/arxiv.2101.03049} have shown remarkable results on $128 \times 128$, $256 \times 256$ and $512 \times 512$ spatial resolutions. Owning to the higher resolution and disentangled latent space of the StyleGAN, multiple works in this domain either use the pretrained StyleGAN generator to construct a video generation framework~\cite{fox2021stylevideogan, DBLP:journals/corr/abs-2201-13433, https://doi.org/10.48550/arxiv.2201.08361} or reformulate the problem by training additional modules on top of StyleGAN and using the video data to train the networks~\cite{stylegan_v, wang2022latent, tian2021a}. Among them is StyleVideoGAN~\cite{fox2021stylevideogan} which is based on the manipulation in $W+$ space of StyleGAN. Related to the pretrained latent space based method, Third Time's the Charm~\cite{DBLP:journals/corr/abs-2201-13433} and Stitch it in Time~\cite{https://doi.org/10.48550/arxiv.2201.08361} use the analysis in the $W$ and $S$ spaces of StyleGAN to edit an embedded video. Note that these methods are concurrent to our method and are only available on arXiv at the time of this submission. In addition, these methods solve a different task than ours and are focused on editing an embedded video in different spaces of StyleGAN. Another set of works like StyleGAN-V~\cite{stylegan_v} and LIA~\cite{wang2022latent} retrain the modified StyleGAN architecture on videos. Note that our method is a latent space based method on StyleGAN3 trained on images that does not require additional video training. LIA is also trained on different datasets than ours and cannot control the individual components of the generated image by deriving information from different videos. Also note that  StyleVideoGAN, Third Time's the Charm, and Stitch it in Time use $W+$ and/or modified weights of the generator~\cite{roich2021pivotal} to embed videos into the latent space. To compare with such methods, we construct a baseline based on such methods in Sec~\ref{sec:results}, where $W+$ space is used to encode the global and local components of the video.

\vspace{-0.3cm}
\section{Method}
\label{sec:method}

Given a reference image $I_{ref}$ and frames of a driving video $D$, our goal is to produce a final sequence of video frames $V=:\{V_j\}$ that enacts a talking head with the identity of 
$I_{ref}$ and pose and expressions, both local and global, from the driving video $D$. Optionally, a co-driving video $CD:=\{ CD_j \}$ may be provided as input. We develop a framework based on these parameters to produce a disentangled representation of a driving video, such that we are able to encode both its global and local properties, and  control them separately to produce an output video $V$.

\begin{figure}[t!]
    \centering
    \includegraphics[width=\linewidth]{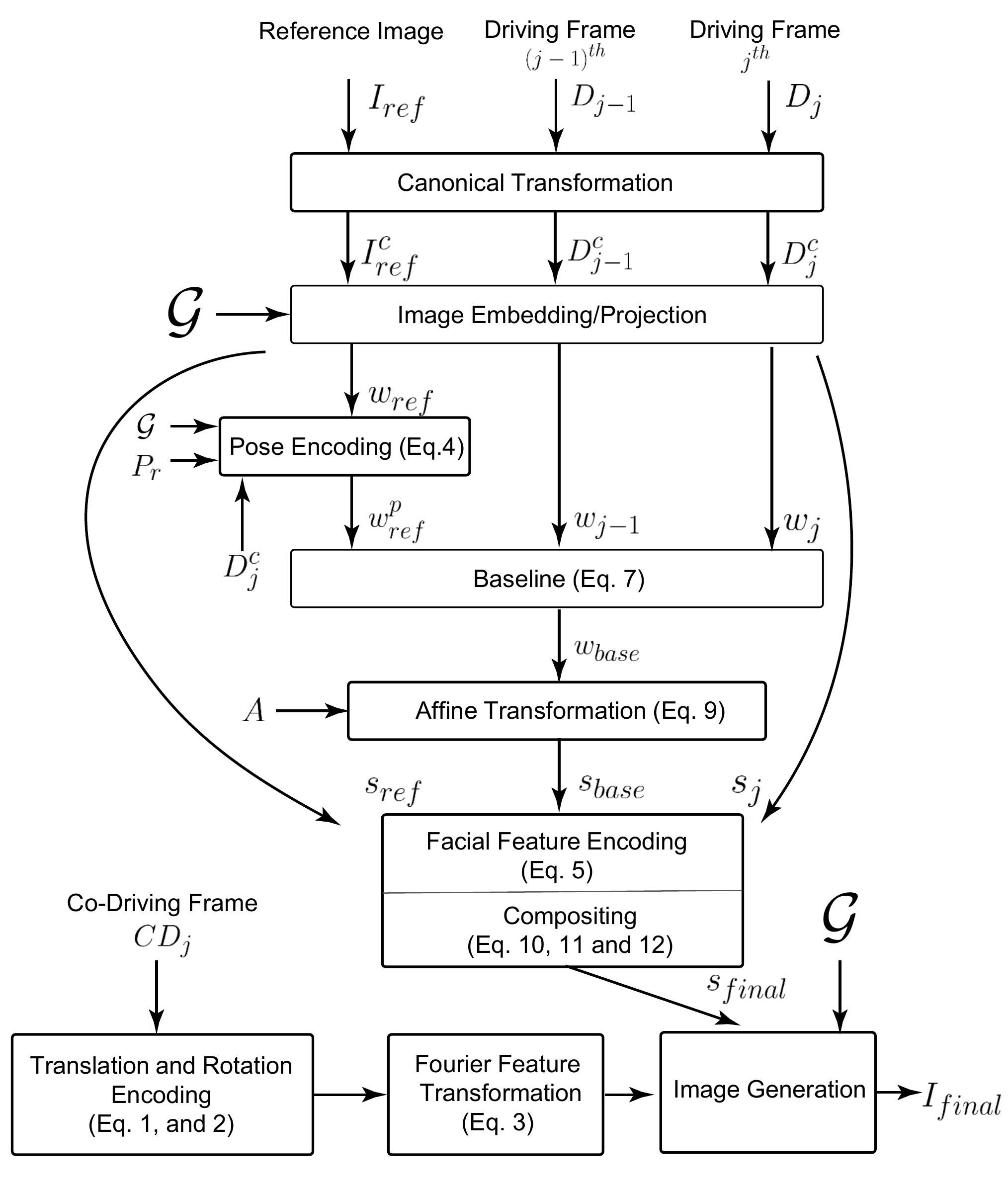}
    \caption{ 
    \textbf{\name pipeline. }
    Flow diagram of our Video2StyleGAN method. Each box represents a local or global encoding and editing module used by our method. See Section~\ref{sec:method} for details. 
    }
    \label{fig:framework}
\end{figure}

\subsection{Overview}

Our framework, see Fig~\ref{fig:framework},  for reenactment of the talking head, using a single (identity) image, is based on the analysis of two spaces associated with StyleGAN3, namely the $W+$ and $S$ spaces. Let $w+ \in W+$ and $s \in S$ be the variables in the respective spaces for any input image. We recall that parameters in the $S$ space are derived from the $w+$ codes using $s := A(w+)$, where $A$ is an affine transformation layer in the StyleGAN3~\cite{karras2021aliasfree}. Let $\gen$ be the pretrained StyleGAN3 generator. In addition to these two latent  spaces, we note that the first layer of the StyleGAN3 $\gen$ produces interpretable Fourier features, parameterized by translation and rotation parameters. We represent this function as $F_f$.

Previous research has shown that the $W+$ space~\cite{abdal2020styleflow} encodes global properties and semantics of an object; while the $S$ space~\cite{wu2020stylespace} is more suited for encoding local variations in a StyleGAN generated image. Based on this observation, we make use of \textit{both} of these spaces in our framework. Specifically, to separately control the local and global variations in our resulting video, we classify such components as local or global editing techniques. Note that some components of the framework includes a combination of both these spaces to achieve a desirable reenactment (Section~\ref{sec:add_enc}). \par

In order to encode a given driving video into the latent space of StyleGAN3, we first project the individual frames of the video into the StyleGAN3 latent space. We use the state-of-the-art projection method, ReStyle~\cite{alaluf2021restyle} to project the canonical frames of the video and the reference image (i.e., after the FFHQ based transformation) into the $W+$ space ($W+ := \{{w+}\in \mathbb{R}^{18 \times 512}\}$) of StyleGAN3. Let the resulting reference image be represented by $I^c_{ref}$ and $w_{ref}$ be the corresponding $w+$ code. In the following subsection, we first  introduce the building blocks of our video editing framework, and then combine them to produce \name, a controllable video generation framework.

\begin{figure}[b!]
    \centering
    \includegraphics[width=\linewidth]{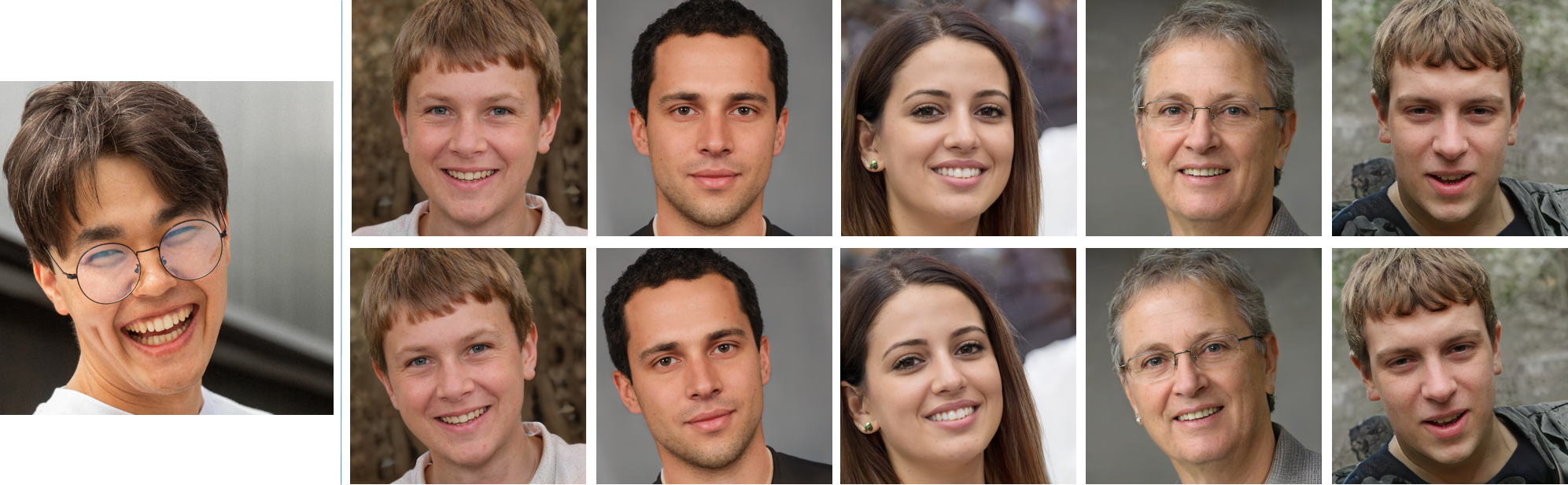}
    \caption{
    \textbf{Canonical transformation. }
    Given a driving video (left) with rotation and translation of a driving frame, our framework can transfer this information to a new reference image. Top row: reference images. Bottom row: transformed images.
    }
    \label{fig:tranform}
\end{figure}
\vspace{-0.3cm}

\subsection{Canonical Transformation}
\label{sec:rot_trans}

Given a video to define the position of the face within a frame, we exploit the translation and rotation invariance property of the StyleGAN3 architecture to encode the rotation and translation of the talking head. We recall that the Fourier features of StyleGAN3~\cite{karras2021aliasfree} can be transformed to produce an equivalent effect on the output image. We define a tuple $(t_x,t_y,r)$, where $t_x$ and $t_y$ are the horizontal and vertical translation parameters, and $r$ is the rotation angle. 
First, in order to determine the translation and rotation changes from the canonical positions present in FFHQ~\cite{karras2019style}, we use a state-of-the-art landmark detector~\cite{face_recog} on each frame of the video to determine the frame-specific $(t_x,t_y,r)$  parameters. For each frame, we compute a vector connecting the average of the positions of the \textit{eye} landmarks and the \textit{mouth} landmarks. We use them to compute the relative angle between the canonical vertical vector and the current face orientation that we use to encode the rotation of the head. Formally, let $e_l$ and $e_r$ be the eye landmarks (left and right, respectively) and $m_l$ be the mouth  landmarks predicted by the landmark detector $L_d$. Then,

\begin{equation*}
    \vec{e} := 0.5 (\mathop{\mathbb{E}} (e_l) + \mathop{\mathbb{E}} (e_r)) 
    \quad \text{and} \quad \vec{v} := \mathop{\mathbb{E}} (m_l) -  \vec{e}
\end{equation*}
and
\begin{equation}
    r := d_{cos}(\vec{u}, \vec{v}), 
\end{equation}
where $\mathbb{E}$ denotes average, $d_{cos}$ is the cosine similarity function, and $\vec{u}$ is the up vector. 
 Similarly, as per the FFHQ transformation, the translation parameters are given by,
 \begin{equation}
    \vec{t} := \vec{e} - \vec{e}' , 
\end{equation}
 where $\vec{e}'$ is the  midpoint of the canonical FFHQ transformed image, and $\vec{t}$ is a column vector representing $t_x$ and $t_y$.
 The transformations can then be performed on the Fourier features $F_f$ to produce the desired rotation and translation effects on a given image, as, 
 \begin{equation}
    F_f'(t_x,t_y,r) := F_f(\tau(t_x,t_y,r))
\end{equation}
 where $\tau$ represents the transformation. See Fig.~\ref{fig:tranform} for visual results.

These transformations~\cite{DBLP:journals/corr/abs-2201-13433}, however, are not smooth across the frames. Hence, in order to smoothen out the anomalies, we apply a convolution operation to this sequence of parameters across the time domain. Empirically, we found a mean filter with a kernel size of $3$ or higher to produce a smooth consistent video after the transformations are applied. Note that these parameters can also be derived from another co-driving video $CD$, other than the driving video $D$, and applied to a given image without affecting the identity. For example, we could apply the steps mentioned in the following section from a first driving video and apply the rotation and translation effects from a  second co-driving video.

     \vspace{-.5in}
 \begin{figure}[h!]
    \centering
    \includegraphics[width=\linewidth]{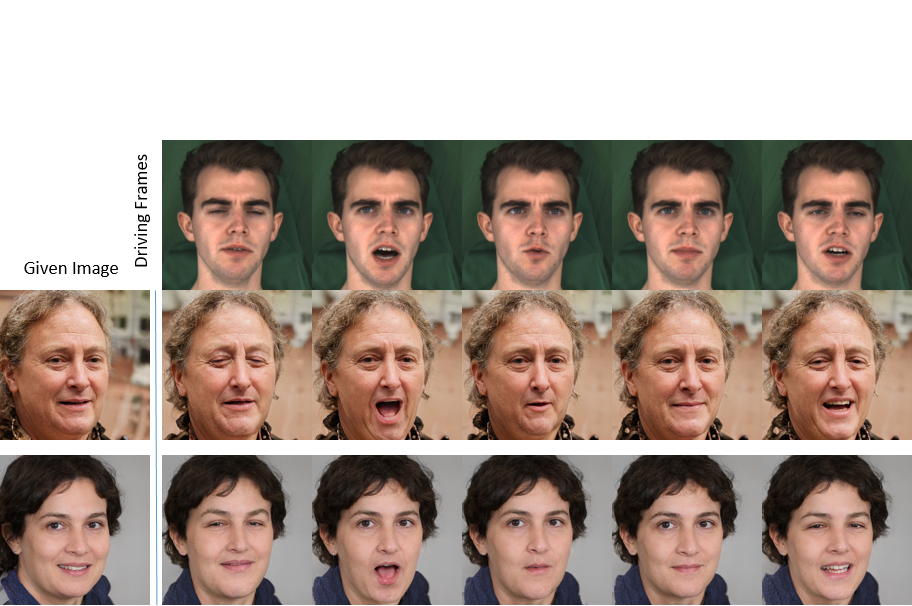}
    \caption{
    \textbf{Editing global pose. }
    Given a reference image and a driving frame, we match the pose and local information. Each row shows an identity (reference image) and edits corresponding to the top row (driving frame). }
    \label{fig:pose}
\end{figure}

\subsection{Global Pose Encoding}
\label{sec:pose}
Next, we encode the pose of the face in the driving video. We consider this also as a global variation of the video editing framework. Pose changes in StyleGAN are largely associated with adding new details --- stretching, squeezing, and transforming the eyes and mouth views to a target position. For this reason, we use the $W+$ space of the StyleGAN3 to encode such global information. Given the hierarchical structure of the StyleGAN3 architecture and semantic understanding of the latent space~\cite{abdal2020styleflow}, we restrict encoding the pose information in the first $8$ (out of $16$) layers of the StyleGAN3 latent space. First, we observe that when we render the given image by transferring the first two layers of the $w+$ code of a driving frame (${w+}\in W+$), it stretches the face area and the eye, however mouth and nose positions remain unchanged making the output face unrealistic. While transferring the first eight layers makes a plausible pose change at the cost of identity loss (see supplementary video).

To circumvent this, we setup an optimization to match the pose information. Specifically, we setup an objective to optimize for pose (i.e., yaw, pitch, and roll) of a given image to match the pose of the driving video. We propose two loss functions to ensure the pose matching and identity preservation, respectively. To ensure pose preservation, we use a pose regression model~\cite{landmark} which, given a valid frame of video, outputs yaw, pitch, and roll. To ensure identity preservation, we apply an additional loss to ensure the features of the face and the identity are preserved. The latter loss ensures that the optimized latents from the first $8$ layers remain as close to the original latents of a given image as possible. Note that for editing real faces from photographs, we use a PTI based method~\cite{roich2021pivotal} to project the images. In this case, there are multiple ways to optimize the code and we observed that optimizing the code with the original generator for pose and then using the PTI trained generator to fill in the details works the best. The optimization is given by:
\begin{equation} 
w^p_{ref} := \argmin_{w_{ref}^{1:8}} \underbrace{L_{mse} ( P_r (G (w_{ref})), P_r (D_j ))}_\text{pose preservation}  + \underbrace{L_1(w_{ref}^{1:8}, w^{p 1:8}_{ref})}_\text{identity preservation},
\end{equation}
where $w_{ref}^{1:8}$ represents the $w+$ code of the reference image for the first eight layers of StyleGAN3, $L_{mse}$ represents the $MSE$ loss, and $P_r$ is the output of the pose regression model~\cite{landmark}.

\begin{figure}[b!]
    \centering
    \includegraphics[width=\linewidth]{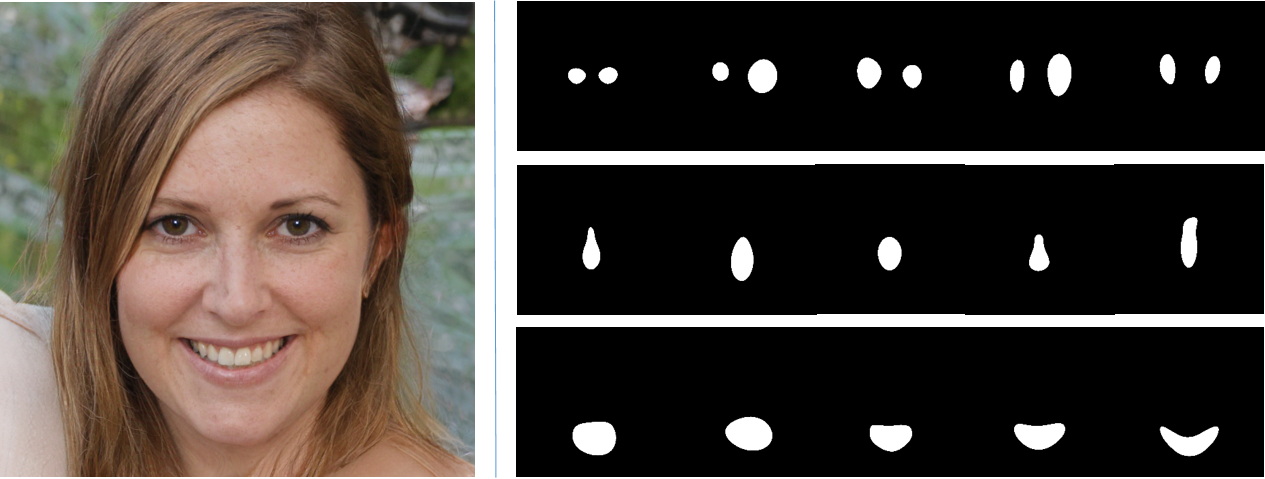}
    \caption{
    \textbf{Extracting local facial features. }
    A given image and some normalized feature maps extracted by the Local Facial Feature Encoding of our framework (see Eq~\ref{eq:iou}). The top row shows maps focusing on the eyes, the middle row shows maps focusing on the nose, and the bottom row shows images focusing on the mouth areas. We restrict latent space optimization to discovered channels responsible for the corresponding edits. }
    \label{fig:activation}
\end{figure}

\begin{figure*}[t]
    \centering
    \includegraphics[width=0.9\linewidth]{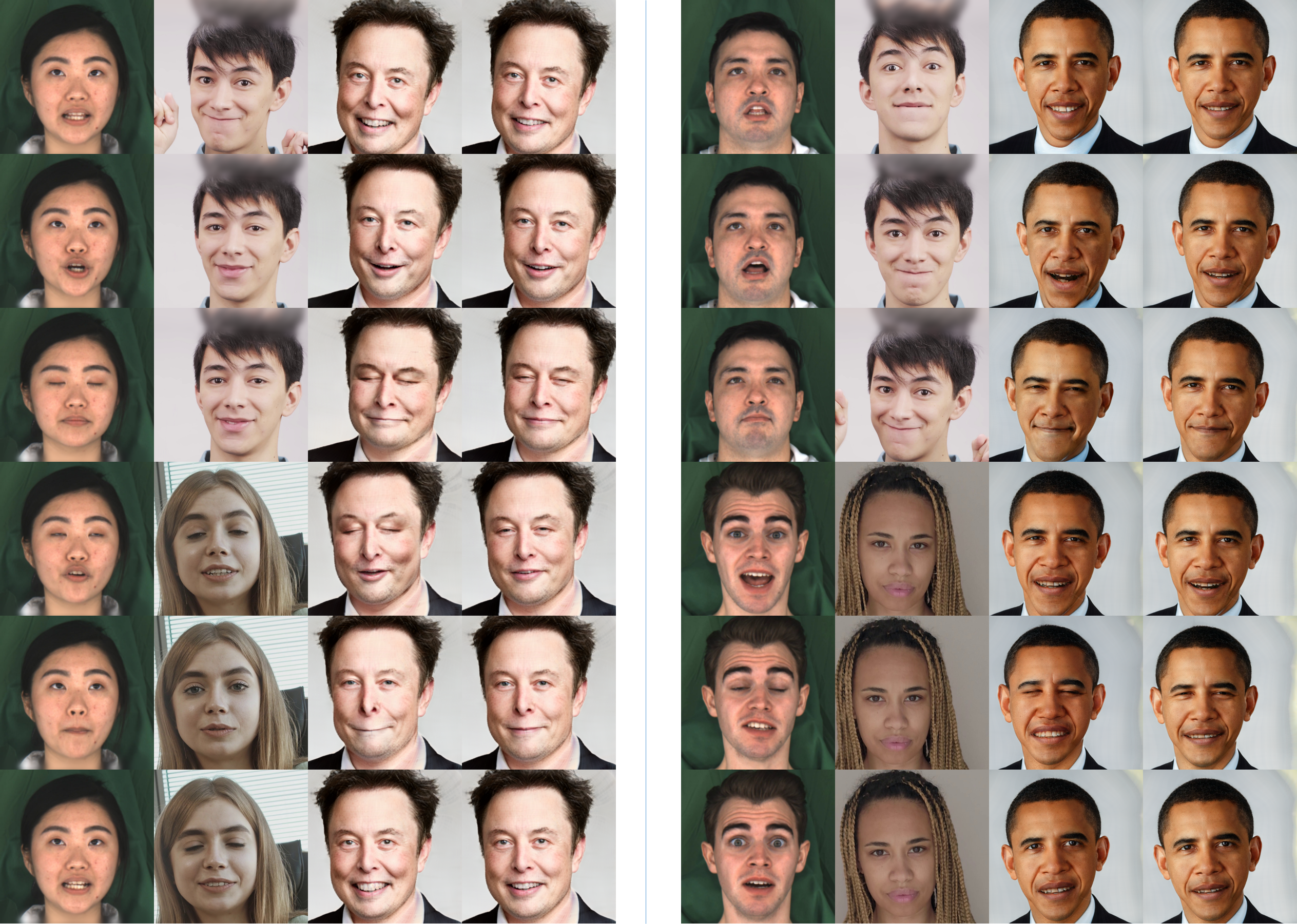}
    \caption{
   \textbf{ Comparison with the baseline. }
    In each sub-figure, the first column shows the driving frames, the second column shows the co-driving frames, the third column shows the results of the baseline method, and the last column shows our results.
    Please see the supplementary video. 
    }
    \label{fig:real_images}
\end{figure*}

In Fig.~\ref{fig:pose}, we show the results of the pose matching from a random frame in the driving video. The figure shows different results of pose changes made to the reference images under a given pose scenario in the driving video.

\subsection{Local Facial Feature Encoding}
\label{sec:enm}

After encoding the global information in a video, i.e., rotation, translation, and pose of the talking head, the next step is to focus on local information. We identified three semantic parts, namely  \textit{eyes}, \textit{mouth}, and \textit{nose}, as being responsible for local changes in a talking head driving video.
Note that often expression changes result in coupled variations across these semantic regions. 
In order to capture the local variances of these semantic parts of the face, we resort to our analysis in the $S$ space of the StyleGAN3 architecture. In order to appreciate the disentangled nature of the $S$ space in StyleGAN3, we encourage the reader to refer to the video of a UI application in the supplementary material showing the local properties of $S$ space and the corresponding feature maps, which can be edited. Just for demonstration, we manually change the $S$ parameter of a given channel of a given layer of StyleGAN3 to observe the desired effect on the final image. We now use this observation for an automatic procedure for manipulating parameters in the $S$ space. 

In order to automatically identify the feature maps and the corresponding $s \in S$ parameters responsible for affecting the motion of the semantic regions, we match the activations in these layers with semantic segmentation regions obtained using a segmentation network. We use a semantic part segmentation network, BiSeNet~\cite{bisenet}, trained on the CELEBA-HQ~\cite{karras2017progressive} dataset, to determine such layers. First, given a set of images and their feature maps extracted from the StyleGAN3, we first compute the segmentation map of the image using BiSeNet. Second, we compute the normalized maps using $\min-\max$ normalization per feature channel of the feature maps. Third, to match the spatial size of these masks, we upsample these masks to match the spatial size of the target mask using bilinear interpolation. In order to convert these normalized features into hard masks, we threshold these maps to be binary. Finally, we compute the $IOU$ scores of the three semantic components derived from the set of images by comparing with these binary masks. 

Let $SegNet$ be the semantic part segmentation network (e.g., BiSeNet), $M_{fg}$ be the semantic component in consideration, $M_{bg}$ be other semantic components including background given by $SegNet(I^c_{ref})$. Let $C_l$ be the feature map at layer $l$ of StyleGAN3 after applying the $\min-\max$ normalization, upsampling and binarization to the map, to produce,  
\begin{eqnarray}
IOU^+ &:=& IOU(M_{fg}, SegNet(C_l)) \quad \text{and} \nonumber \\
IOU^- &:=& IOU(M_{bg}, SegNet(C_l)).
\end{eqnarray}
Based on both the positive $IOU^+$ (eye, nose, and mouth) and negative $IOU^-$(background and components excluding the given semantic part) $IOU$-s, we select a subset of these maps ($\mathcal{X}_m := {\{x}\in \mathbb{R}^{1024^2}\}$) and the corresponding $s$ parameters ($\mathcal{X}_s := {\{x}\in \mathbb{R}\}$) based on thresholding to be our local model for the manipulation of the semantic parts. Formally,
\begin{eqnarray}
C_l \in \mathcal{X}_m,  \ \text{if}  \ IOU^+ 	\geq t_{fg} \ \text{and} \ IOU^-\geq t_{bg} 
\label{eq:iou}
\end{eqnarray}
where $t_{fg}$ and $t_{bg}$ are the thresholds. Note that $\mathcal{X}_s \subset S$.
In Fig.~\ref{fig:activation}, we show some examples of the extracted feature maps in $\mathcal{X}_m$ focusing on only a specific semantic part of the face.

\begin{figure}[t]
    \centering
    \includegraphics[width=\linewidth]{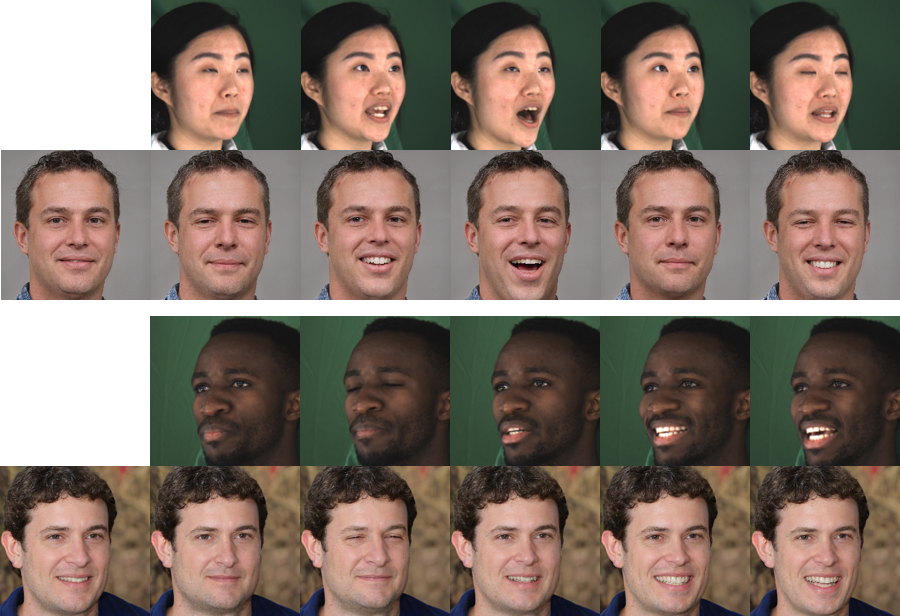}
    \caption{
    \textbf{Fine-grained local control. }
    Local information transfer without the global changes like pose. In each sub-figure, top row represents driving frames and the bottom row shows a reference image and local edits. }
    \label{fig:wo_pose}
\end{figure}

\subsection{Baseline Video Editing}
\label{sec:add_enc}

In our experiments, we found that it is sufficient to simply encode the above global and local components to perform realistic video editing using the StyleGAN3 generator. We observe that even though the $w+$ code of the projected driving video can encode non-semantic components, which cannot be directly used for the video editing, it carries other important information that is lost when shifting to the $S$ space analysis described above. Specifically, in Fig.~\ref{fig:real_images}, we simply compute the difference vectors in the consecutive frames of the driving video, and apply these transformations to the given latent representing a given image. Formally, 
\begin{equation}
w_{base} = w^p_{ref} + (w_{j-1} - w_j)
\label{eq:w_set}
\end{equation}
where $w_{j-1}$ is the $w+$ code corresponding to $D_{j-1}$ and $w_j$ is the $w+$ code corresponding to $D_j$ of the driving video. \par

Notice the artifacts and loss in the identity of the person using such a naive technique for the video editing (see supplementary video). We regard this as a baseline for our method. 
Nevertheless, we observe that the $w+$ code, despite having some undesirable effects, captures some additional semantics essential for making the motion of the face consistent with the driving video. For example, we observe non-local effects such as stretching and squeezing of the cheek during the movements in the mouth, eye regions, and in the chin. Such coupling between the (semantic) parts cannot be captured by only a local analysis. 

\subsection{\name Method}

We now combine all these components to present our  \name (see Figure~\ref{fig:framework}). Note that in our framework, components can be separately controlled and driven by multiple videos (see Section~\ref{sec:results}). 

First, we canonicalize the input video(s) by computing the rotation and translation parameters of the driving or a co-driving video using the Translation and Rotation Encoding (Section~\ref{sec:rot_trans}), and use the extracted transforms on the given image. Alternatively, we can omit these changes to stay faithful to the original parameters in a given image.
    
Then, we perform pose changes via the driving or co-driving video using Pose Encoding (Section~\ref{sec:pose}). Again, we  can omit such changes, i.e., use the pose of the given image without matching it to a driving frame.
    
Finally, we combine the components of the $W+$ space (Section~\ref{sec:enm}) and $S$ space analysis (Section~\ref{sec:add_enc}) to achieve fine-grained control over the video generation. In particular, the degree of local changes can be modified by the $s^p_{ref} \in \mathcal{X}_s$ and combined with the $W+$ space analysis based method. In practice, we identify the $W+$ code layers $3-7$ to produce best results when combined with the $S$ space. Let $\mathcal{X}_{orig} := \{{x}\in \mathbb{R}^{512}\}$ be the original $w+$ encoding of the given image using layers $3-7$.  Similarly, we denote another set of $w+$ codes obtained from Eq~\ref{eq:w_set} as $\mathcal{X}_w := \{{x}\in \mathbb{R}^{512}\}$. Let $A_l$ be the affine function of layer $l$ of the StyleGAN3 generator. Then, we compute the original $s$ parameters as:
    \begin{equation}
    \mathcal{X}_{origs} :=  \bigcup\limits_{i=3}^{7} A_l(w_l),
    \end{equation}
    where $w_l$ $\in$ $\mathcal{X}_{orig}$. Similarly, we compute the $s$ parameter contribution of edited frames as :
     \begin{equation}
    \mathcal{X}_{ws} :=  \bigcup\limits_{i=3}^{7} A_l(w'_l),
    \end{equation}
    where $w'_l$ $\in$ $\mathcal{X}_w$.\par
    
    We encode the local changes by adding the computed $s$ parameters of the given image ($\mathcal{X}_s$, see Section~\ref{sec:pose}) and a given frame of the driving video ($\mathcal{X}_{sd}$). Formally, let $s^p_{ref}$ $\in$ $\mathcal{X}_s$ and let $s^p_j$ $\in$ $\mathcal{X}_{sd}$ be the $s$ parameters in the given sets, then the operation is given by:
    \begin{equation}
    s_{local} = \alpha {s^p_{ref}} + \beta {s^p_{j}}.
    \end{equation}
    Combining the two, we define the manipulations: 
    \begin{equation}
    s_{final} = s_{local} + \gamma s^{p}_{base},  
    \end{equation}
    where $s^{p}_{base}$ $\in$ $\mathcal{X}_{ws}$, such that it matches the $s$ parameter position computed in set $\mathcal{X}_s$. For other $s$ parameters:
    \begin{equation}
    s_{final} = \zeta s^{q}_{ref} + (1-\zeta) s^{q}_{base}
    \end{equation}
    where $s^{q}_{ref} \in $ $\mathcal{X}_{origs}$ and  $s^{q}_{base}\in \mathcal{X}_{ws}$. Note that $\alpha, \beta, \gamma,  \zeta $ can be controlled separately to produce a desirable animation.

\section{Results}
\label{sec:results}

\subsection{Training and Implementation details}

We use an Nvidia A100 GPU for the experiments. We use the R-Config model of the StyleGAN3 for the inference. Starting from a pose in the reference frame, Pose Encoding takes under 1 minute to converge. We set the yaw, pitch, and roll loss weights to $2$ and the identity preservation loss weight to $0.04$. As a default setting, we set $\alpha = -1$, $\beta = 1$, $\gamma = 1$ and $\zeta = 0.5$.

\subsection{Baseline}

As a baseline method, we resort to Eq.~\ref{eq:w_set} as method to make consecutive edits to the $w+$ code of the embedded video. Note that this method is widely used by Gan-based image editing methods like InterfaceGAN~\cite{shen2020interfacegan} and GANSpace~\cite{harkonen2020ganspace}. More specifically, current video editing works on arXiv~\cite{DBLP:journals/corr/abs-2201-13433,https://doi.org/10.48550/arxiv.2201.08361} use the videos embedded in the $W+$ space and/or weights of the generator~\cite{roich2021pivotal} to do further editing. We apply the same approach to modify a single image and generate a video using the driving and the co-driving frames. In Fig~\ref{fig:real_images}, third column in each sub-figure shows the result of the baseline method on two different identities.

\subsection{Metrics}

We use three metrics to evaluate the keypoints, identity preservation, and the quality of the frames in the resulting video. We also check the consistency of these metrics on the resulting videos (Sec~\ref{sec:quant}) by encoding a reverse driving video. These metrics are:

\begin{enumerate}
    \item \textbf{Keypoint Distance ($\Delta K$):} In order to measure the target edits made to the resulting video we use the Mean Squared Error between the keypoints of the driving video and the resulting video. The keypoint detector~\cite{landmark} predicts $68$ keypoints given an image. We average the errors for these $68$ keypoints.
    \item \textbf{Identity distance $ID$:} We use a state-of-the-art Face recognition model~\cite{face_recog} to compute the facial embeddings of the given reference image and the frames of the resulting video. We compute $L2$ norm of these embeddings and take an average across all frames of the video.
    
    \item \textbf{Fréchet Inception Distance (FID) :} FID is used to measure the distance between a given distribution of images to a generated one. In the context of video editing, we use FID as a metric that computes the quality and consistency of the edits made to the given reference frames. This measures how much does the distribution of the resulting frames change under different edits (e.g., by reversing the driving video).
\end{enumerate}

\subsection{Qualitative Comparison}

In order to visualize the quality of the resulting video, in Figure~\ref{fig:teaser}, we show the results of our Video2StyleGAN method on different identities. Note that here we first match the pose of the given identity image to a driving frame and then we apply the local and global edits including the rotation and translation derived from a co-driving video. Notice the quality of the identity preservation across different editing scenarios. To compare our method with the baseline, in Fig~\ref{fig:real_images}, we show the results of the editing and transformations. For embedding a real image, we use the Restyle method to produce an embedding and further optimize the generator using PTI~\cite{roich2021pivotal} by initializing with the computed Restyle embedding. Notice that the baseline approach tends to change different features like skin color and produces noticeable artifacts. In comparison, our method is able to preserve the identity of the person and successfully transfer the edits from the driving and the co-driving video. In order to show that our method works when the pose of the reference image does not match the driving frame, in Fig~\ref{fig:wo_pose}, we show the transfer of the local information from the driving frames to a reference image. Notice the quality of edits and identity preservation in these case. Please refer to the supplementary video. 

\subsection{Quantitative Comparison}
\label{sec:quant}

In order to compute the metrics on frames of the video produced by our Video2StyleGAN method and compare with the baseline, we use $5$ identities to produce a video of $160$ frames using the video collected from the MEAD~\cite{kaisiyuan2020mead} dataset. To test the consistency of the methods, in addition to computing the edits in the forward direction, we reverse the driving video and compute the edits using this reverse driving video. A consistent method should produce similar edits starting from a reference image, such that the identity, keypoints and the quality of the edits are preserved. In Table~\ref{tab:keypoint}, we compute the three metrics $\Delta K$, $ID$ and FID using both the driving as well as the reverse driving video. In case of Keypoint Distance ($\Delta D$) (computed for both $x$ and $y$ positions of the keypoints) notice that our method beats the baseline method in both scenarios showing that our method is both better at matching the keypoints as well as consistent across the driving video direction. Similarly, Identity Distance ($ID$) in our case is much better than the baseline and is consistent across the driving video direction. Finally, in order to compute the quality and consistency of the edits, we measure the FID score between the frames produced by a driving video and its reverse version. The table shows that our results are better in this case as well. This indicates that our method is able to produce consistent quality of images across different identities.

\begin{table}%
\caption{
\textbf{Quantitative evaluation. }
Table showing the evaluation of the Keypoint Distance ($\Delta K$), Identity Distance $ID$, and FID scores on the baseline method and ours. Lower is better for all the metrics. Here superscript $f$ represents a metric evaluated on a forward driving video and $r$ represents the scores on the reverse driving video. Subscripts $x$ and $y$ represent the evaluation of the two 2D coordinates of the keypoints. See text for details. }
\label{tab:keypoint}
\begin{minipage}{\columnwidth}
\begin{center}
\begin{tabular}{rllllllll}
  \toprule
  Method  & $ \Delta K^{f}_{x} $   & $ \Delta K^{f}_{y} $  & $ \Delta K^{r}_{x} $ & $ \Delta K^{r}_{y}  $ & $  ID^{f} $   & $  ID^{r} $ & FID\\ \midrule
Baseline   & 0.19  &0.40 & 0.26  &\textbf{0.28} & 0.46  &0.51 &17.34   \\

 Ours    & \textbf{0.17}  & \textbf{0.36} & \textbf{0.20}  &0.29   & \textbf{0.24}  & \textbf{0.23} &\textbf{6.15} \\

  \bottomrule
\end{tabular} 
\end{center}
\end{minipage}
\end{table}%

\section{Conclusions}
We introduced a framework for fine-grained control for manipulating a single image using the StyleGAN3 generator. In particular, the framework is useful to edit a single image given a driving video. This problem is very challenging, because existing methods either strongly overfit or underfit the driving video. Our experiments yield qualitative results in the accompanying video and quantitative results using three different metrics to demonstrate a clear improvement over the current state of the art, including recent arXiv papers.
Our work also has some limitations. Our method is image-based, and we do not reconstruct a complete 3D model. This means we trade 3D and viewpoint consistency for a higher visual quality of details. Further, our algorithm currently only considers face models. Being able to handle complete human bodies would require further extensions.
In future work, we would like to explore video editing using other generative models, such as auto-regressive transformer and diffusion. We also propose text driven video editing as a possible direction for future work.

\begin{acks}

We would like to thank Visual Computing Center (VCC), KAUST for the support, gifts from Adobe Research, and the UCL AI Centre. 

\end{acks}

\bibliographystyle{ACM-Reference-Format}
\bibliography{references}

\end{document}